\documentclass[11pt]{article}

\usepackage[utf8]{inputenc}
\usepackage[T1]{fontenc}
\usepackage{lmodern}
\usepackage[a4paper,margin=1in]{geometry}
\usepackage{amsmath}
\usepackage{graphicx}
\usepackage{booktabs}
\usepackage{tabularx}
\usepackage{array}
\usepackage{float}
\usepackage[authoryear,round]{natbib}
\usepackage{authblk}
\usepackage{microtype}
\usepackage[hidelinks]{hyperref}
\usepackage{caption}

\graphicspath{{figures/}}
\newcolumntype{L}{>{\raggedright\arraybackslash}X}

\title{Nepali Currency Denomination Classification Using Convolutional Neural Networks (CNNs) and MobileNetV2: An Experimental Comparison}

\author[1]{Sakar Lamichhane\thanks{\texttt{sakar.sk47@gmail.com}}}
\author[2]{Sandesh Bhatta\thanks{\texttt{sbhatta@albany.edu}}}
\author[3]{Aayush Adhikari\thanks{\texttt{ayushadhikari3466@gmail.com}}}
\author[4]{Aaron Sapkota\thanks{\texttt{aaronsapkotanasa@gmail.com}}}
\affil[1]{Islington College, Kathmandu, Nepal / London Metropolitan University, London, United Kingdom}
\affil[2]{Department of Computer Science, University at Albany, State University of New York (SUNY), Albany, NY, USA}
\affil[3]{Department of Computer Science \& Engineering, C.V. Raman Global University, Bidyanagar, Mahura, Bhubaneswar 752054, India}
\affil[4]{University of Technology Sydney, Sydney, New South Wales, Australia}
\date{}

\begin{document}

\maketitle

\begin{abstract}
Automated currency denomination recognition is a computer vision task with applications in cash handling, financial technology, and assistive systems. This study classifies seven Nepali currency denominations (Rs. 5, 10, 20, 50, 100, 500, and 1,000) using three configurations: a conventional convolutional neural network (CNN), MobileNetV2 with transfer learning, and MobileNetV2 with hyperparameter tuning. A compiled dataset of 16,249 images was used, comprising 10,500 training, 3,500 validation, and 2,249 independent test images. Images were resized to 128 x 128 pixels, normalized, and augmented to vary orientation, scale, illumination, and appearance. Models were evaluated using accuracy, precision, recall, F1-score, and precision-recall and ROC analysis. The hyperparameter-tuned MobileNetV2 achieved the highest validation accuracy (95\%), with precision, recall, and F1-score of 0.93, 0.91, and 0.92, but reached only 84\% accuracy on the held-out test set. MobileNetV2 with transfer learning but no additional tuning achieved 89\% test accuracy, and the conventional CNN achieved 85\%. Transfer learning therefore provided a competitive approach under the evaluated conditions, while higher validation accuracy from tuning did not yield higher test accuracy. To demonstrate practical deployment, a Flask-based web prototype using TensorFlow Lite was developed, supporting image upload and live webcam recognition with both visual and voice output. The study shows the feasibility of combining lightweight deep learning with a voice-assisted web interface for Nepali currency recognition, and highlights the need for more diverse datasets, greater robustness under uncontrolled imaging conditions, and broader real-world evaluation.
\end{abstract}


\noindent\textbf{ACM classes:} I.5.4 Pattern Recognition -- Applications; I.2.10 Vision and Scene Understanding.

\section{Introduction}

Currency denomination recognition is a computer vision problem where an image of a banknote is used to identify its monetary value. An automated system for denomination recognition can reduce the need for manual identification. It can also be useful in areas such as financial services and assistive technologies. In countries where cash is still commonly used for everyday transactions, having a reliable system for recognizing local banknotes is especially useful.

Deep learning methods, particularly convolutional neural networks (CNNs), are widely used for image classification because they can learn hierarchical visual representations directly from image data. CNNs can capture patterns related to colour, texture, shape, and other visual characteristics, making them suitable for currency denomination classification. However, visually similar denominations may remain difficult to distinguish, particularly when images contain variations in orientation, lighting, background, or banknote condition \citep{yamashita2018convolutional}. Currency recognition also has an important accessibility aspect. A system that provides its prediction through speech can be useful for users who may not be able to rely entirely on visual information. Voice feedback gives users another way to identify the denomination of a banknote and can make the recognition system more accessible. In this study, voice output is included in the web-based prototype. It is designed to support the visual classification result rather than replace it, so users can receive the predicted denomination both visually and through audio.

The main contribution of this study is the focused evaluation of three CNN-based configurations for seven-class Nepali currency denomination classification. It also includes hyperparameter tuning and a practical web prototype that supports both image upload and live webcam recognition, with the predicted denomination provided through visual and voice feedback. In addition, the study examines class-level behaviour using precision-recall and ROC analysis and discusses the difference between validation performance and performance on the held-out test set.

\subsection{Research Gap}

Previous studies have shown that CNNs, transfer learning, and lightweight architectures can be useful for currency recognition. However, their performance can vary considerably depending on the type of currency, dataset composition, image acquisition conditions, preprocessing methods, and evaluation procedures. Existing studies have largely focused on Indian, Bangladeshi, and global currency classification \citep{das2023utilizing}, and therefore the reported results cannot be directly assumed to generalize to Nepali banknotes, which have their own visual characteristics and security features. In addition, the limited availability of standardized and diverse Nepali currency datasets presents a challenge for developing and evaluating models specifically for this currency.

Earlier research has also investigated Nepali currency recognition using CNNs and transfer learning \citep{shrestha2018cash}. However, there remains a need for a focused comparison of different deep learning configurations under a common seven-class Nepali currency classification task. The present study addresses this by comparing a conventional CNN, MobileNetV2 with transfer learning, and a hyperparameter-tuned MobileNetV2, while also examining class-level performance and classification errors. The resulting model is further integrated into a web-based prototype supporting both image upload and live webcam recognition, with predicted denominations presented through visual and voice output. The voice-assisted functionality is demonstrated as part of the prototype rather than evaluated through a formal accessibility or user study.

Therefore, the research gap addressed by this study is the systematic evaluation and comparison of CNN-based approaches for seven-class Nepali currency denomination classification, together with class-level error analysis and a practical web-based demonstration incorporating visual and voice-assisted recognition.

\subsection{Aim and Research Questions}

The aim of the study is to evaluate CNN-based approaches for Nepali currency denomination classification and to investigate whether transfer learning and hyperparameter tuning can improve classification performance. The study addresses two questions:

\begin{description}
  \item[Research question 1:] How effectively can CNN-based models classify seven Nepali currency denominations under the defined experimental conditions?
  \item[Research question 2:] What classification errors and performance limitations are observed across the seven denominations in the evaluated dataset?
\end{description}

\section{Literature Review}

Deep learning-based currency recognition has been studied using conventional CNNs, transfer learning architectures, lightweight models, and hybrid approaches. The literature shows that classification performance can vary depending on the characteristics of the currency, dataset composition, preprocessing methods, and evaluation procedures. Advances in deep learning, especially in Convolutional Neural Networks (CNNs), show significant potential for currency recognition tasks, as CNNs have emerged as a form of image classification problem due to their ability to automatically learn and extract hierarchical features from images \citep{yamashita2018convolutional}.

Researchers have studied currency classification across different national currencies, with transfer learning and lightweight architectures often used to reduce training requirements and support practical deployment. The literature also includes separate studies on counterfeit detection and assistive currency recognition systems. Although these areas are related to denomination recognition, they should be considered separately because they involve different target classes and evaluation objectives. Recent studies have used CNN architecture to classify and recognize fake and genuine banknotes, while assistive systems have combined CNNs with voice feedback for visually impaired users \citep{thomas2021banknote}.

The focus on deep learning has also encouraged lightweight and hybrid architectures. Data preprocessing, augmentation, and model optimization are important because image distortion and variation can negatively affect recognition results \citep{zhang2019overview}. Hybrid approaches combine CNN feature extraction with other classifiers or detection models, while transfer learning reduces training requirements by reusing features learned from large datasets.

\begin{table}[H]
\centering
\small
\caption{Selected related studies in currency recognition and classification.}
\label{tab:related}
\begin{tabularx}{\textwidth}{@{}l >{\raggedright\arraybackslash}p{3.2cm} >{\raggedright\arraybackslash}p{1.8cm} L@{}}
\toprule
\textbf{Study} & \textbf{Method} & \textbf{Reported acc.} & \textbf{Reported observation} \\
\midrule
\citet{mehta2024forgery} & CNN-SVM hybrid & 94\%--97\% & Reported high classification performance; evaluated on non-Nepali currency. \\
\citet{swami2022indian} & CNN + transfer learning & 98\% & Robust Indian currency classification; noisy conditions unexplored. \\
\citet{reddy2023automated} & CNN + transfer learning & $>$98\% & Portable solution; counterfeit detection not included. \\
\citet{shoumik2022bangladeshi} & Lightweight CNN + transfer learning & 99.97\% & Reported low computational cost; dataset diversity was limited. \\
\citet{kazi2024classification} & Minimalist CNN & 70.47\% & Cross-dataset evaluation highlights generalization difficulty. \\
\citet{nahid2024feature} & YOLOv8 + ResNet-50 & 99.83\% & Reported high accuracy; performance depended on image quality. \\
\bottomrule
\end{tabularx}
\end{table}

\subsection{Related Studies}

\citet{mehta2024forgery} investigated the use of Convolutional Neural Networks and Support Vector Machines in combination to detect counterfeit currency. The CNN component was used for image feature extraction and the SVM classifier for final classification. The reported accuracy ranged from 94\% to 97\%, with balanced precision, recall, and F1-score performance.

\citet{swami2022indian} used CNN and transfer-learning techniques for Indian currency classification and reported 98\% accuracy. The study emphasized automated banknote classification for financial transactions, but performance under noisy conditions was not extensively explored.

\citet{reddy2023automated} presented an automated currency classification system designed to help visually impaired users. Transfer learning and MobileNetV2 were used to create a lightweight model suitable for portable and web-based systems. The dissertation notes that the system did not include counterfeit detection.

\citet{shoumik2022bangladeshi} explored lightweight CNN architectures and transfer learning for Bangladeshi paper currency recognition. Data augmentation techniques such as flipping, rotation, and noise addition were used, and the study reported 99.97\% accuracy. Its principal strength was computational efficiency for resource-constrained deployment.

\citet{kazi2024classification} examined Bangladeshi currency classification using a minimalist CNN optimized for cross-dataset recognition. The reported cross-dataset accuracy was 70.47\%, illustrating that dataset variability and environmental changes can substantially affect recognition performance.

\citet{nahid2024feature} combined YOLOv8 for object detection and ResNet-50 for classification of Bangladeshi currency. The study reported 99.83\% accuracy and demonstrated the benefit of combining fast detection with deep feature extraction.

The reviewed studies show that high classification accuracy can be achieved on specific currency datasets. However, the reported results are not directly comparable because the currencies, dataset sizes, image acquisition conditions, model architectures, and evaluation procedures differ. The reported accuracies are therefore presented as contextual information rather than as a common benchmark. The cross-dataset results reported by \citet{kazi2024classification} further show how variations between datasets can affect recognition performance.

Therefore, the present study does not aim to outperform all previously reported currency recognition results. Instead, it evaluates three model configurations under a common Nepali currency classification task and examines their resulting class-level behaviour.

\subsection{Research Gap from the Literature}

The literature highlights significant progress in currency recognition through the use of CNNs, transfer learning, lightweight architectures, and hybrid models. However, direct comparison between existing studies remains difficult because they use different currencies, datasets, image acquisition conditions, preprocessing methods, and evaluation protocols. These differences can influence the reported classification performance and make it difficult to determine how different approaches perform under the same experimental conditions.

The present study addresses this issue through a controlled comparison of three model configurations: a baseline CNN, MobileNetV2 with transfer learning, and a hyperparameter-tuned MobileNetV2 model. All three configurations are evaluated on the same seven-class Nepali currency classification task, allowing their performance to be compared under consistent conditions. In addition to overall classification metrics, the study examines class-specific behaviour to identify differences in how the models recognize individual denominations. The resulting classifier is also demonstrated through a web-based prototype that includes voice-assisted output.

Counterfeit detection is treated as a separate direction for future research. Although it is related to currency recognition, it requires genuine and counterfeit examples and a dedicated evaluation protocol. Therefore, counterfeit detection is outside the scope of the present classification experiments.

\section{Theoretical Background}

\subsection{Convolutional Neural Networks and MobileNetV2}

Convolutional Neural Networks (CNNs) are widely used for image classification because their convolutional layers can learn spatial feature representations directly from input images. These layers can identify visual patterns such as edges, shapes, textures, and other features that are useful for distinguishing between image classes. Pooling operations can then reduce the spatial dimensions of the intermediate feature representations, which can help reduce computational requirements. Finally, fully connected layers can use the learned features to assign the input image to the target classes.

MobileNetV2 is a lightweight CNN architecture designed to reduce computational requirements while maintaining effective feature extraction. It uses depthwise separable convolutions and inverted residual structures to make the network more computationally efficient. These characteristics make MobileNetV2 suitable for applications where computational resources or deployment requirements may be constrained.

In this study, the conventional CNN is used as a baseline model for comparison. MobileNetV2 is then used to investigate the effect of transfer learning on Nepali currency denomination classification. This allows the performance of the lightweight transfer-learning approach to be compared with that of the baseline CNN under the same classification task.

\subsection{Transfer Learning and Data Augmentation}

Transfer learning is a machine learning technique where a pre-trained model is adapted to a new task by fine-tuning its parameters on a domain-specific dataset. In currency recognition, transfer learning allows pre-trained models to leverage features learned from large-scale datasets and apply them to banknote classification. It allows a model that has been trained on a large image dataset to be used as a starting point for a different classification task. Instead of training the model entirely from the beginning, the knowledge learned from the pre-trained model can be adapted to the new dataset. In this study, pre-trained MobileNetV2 weights were adapted for the classification of seven Nepali currency classes.

Data augmentation was applied during training to introduce greater variation in the input images. The implementation described in the original experimental work included rotation, scaling, horizontal and vertical flipping, brightness adjustment, and zooming. These transformations were used to expose the models to variations in image orientation, scale, illumination, and overall appearance during training. This was intended to improve the models' ability to handle variations in the input images.

\subsection{Evaluation Metrics}

The models were evaluated using accuracy, precision, recall, F1-score, confusion analysis, precision-recall curves, and ROC-AUC analysis. Accuracy represents the proportion of samples that are correctly classified. Precision measures the proportion of positive predictions for a class that are correct, while recall measures the proportion of actual samples belonging to a class that are correctly identified. The F1-score provides a combined measure of precision and recall.

Confusion analysis provides class-level information about correctly classified and misclassified samples, helping to identify which denominations are more difficult for the models to distinguish. Precision-recall curves and ROC curves provide additional views of classification behaviour across different decision thresholds. ROC-AUC summarizes the model's ability to distinguish between classes across these thresholds.

\section{Research Methodology}

The study follows an experimental design in which three CNN-based model configurations are trained and evaluated on the same seven-class Nepali currency classification task. Using the same classification task provides a consistent basis for comparing the different model configurations. The experimental workflow consists of dataset preparation, preprocessing and data augmentation, model development, hyperparameter tuning, model training, independent test evaluation, and demonstration through a web-based prototype. This workflow provides a consistent basis for evaluating the three model configurations and subsequently demonstrating the classifier through a web-based prototype.

\subsection{Data Collection and Preprocessing}

The dataset used in this study was compiled from publicly available Nepali currency image resources and supplemented with collected images. The dataset contains images representing seven Nepali currency denominations: Rs.~5, Rs.~10, Rs.~20, Rs.~50, Rs.~100, Rs.~500, and Rs.~1,000. The images were collected from publicly available datasets, online repositories, and self-collected images captured using high-resolution cameras. The dataset includes variations in lighting conditions, backgrounds, viewing angles, and banknote appearance, providing different image conditions for model training and evaluation.

All images were resized to $128 \times 128$ pixels, and their pixel values were normalized to a range of 0 to 1. Data augmentation was applied during model training to introduce additional variation into the training images. The implementation records rotation, scaling, horizontal and vertical flipping, brightness adjustment, and zooming as augmentation operations.

The dataset was divided into training, validation, and test subsets. The training and validation subsets were balanced across the seven denominations, whereas the held-out test subset contained unequal class counts. This difference in class distribution was retained for the independent test evaluation.

\subsection{Dataset Composition}

\begin{table}[H]
\centering
\caption{Composition of the training, validation, and test datasets.}
\label{tab:composition}
\begin{tabular}{@{}lll@{}}
\toprule
\textbf{Subset} & \textbf{Images per class} & \textbf{Total} \\
\midrule
Training   & 1,500 & 10,500 \\
Validation & 500   & 3,500 \\
Test       & Unequal across classes & 2,249 \\
Overall    & --    & 16,249 \\
\bottomrule
\end{tabular}
\end{table}

\begin{table}[H]
\centering
\caption{Class distribution in the test dataset.}
\label{tab:testdist}
\begin{tabular}{@{}lr@{}}
\toprule
\textbf{Denomination} & \textbf{Number of test images} \\
\midrule
Rs.~5     & 130 \\
Rs.~10    & 513 \\
Rs.~20    & 485 \\
Rs.~50    & 149 \\
Rs.~100   & 661 \\
Rs.~500   & 151 \\
Rs.~1,000 & 160 \\
\midrule
\textbf{Total} & \textbf{2,249} \\
\bottomrule
\end{tabular}
\end{table}

\begin{figure}[H]
\centering
\includegraphics[width=0.6\textwidth]{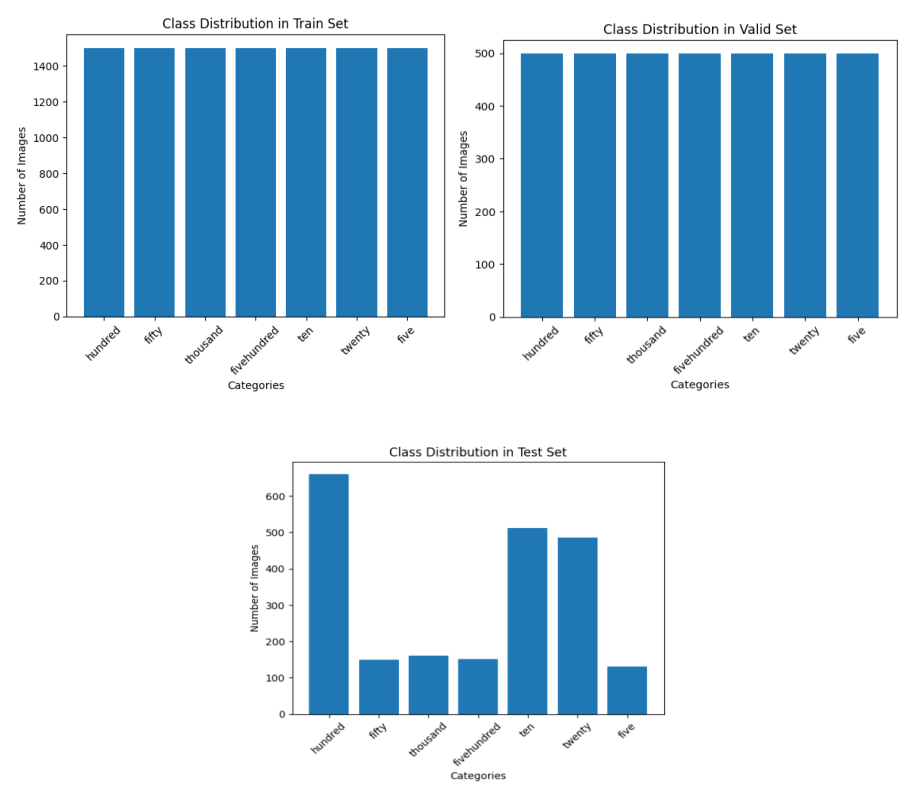}
\caption{Distribution of images across the training, validation, and test subsets.}
\label{fig:split}
\end{figure}

\subsection{Model Development}

Three model configurations were implemented for the seven-class classification task. The first was a baseline CNN trained from scratch using convolutional, pooling, and dense layers. The second used a pre-trained MobileNetV2 architecture with transfer learning. The third extended the MobileNetV2 approach through hyperparameter tuning. This setup allowed the three configurations to be compared under the same classification task.

Hyperparameter tuning was performed using Keras Tuner's random-search approach. The documented search considered learning rates between $1 \times 10^{-3}$ and $1 \times 10^{-5}$, batch sizes of 16, 32, and 64, dropout values between 0.2 and 0.5, and dense-layer sizes between 128 and 512 neurons. These parameters were varied to examine their effect on model performance. The selected configuration was determined based on validation accuracy.

\subsection{Training Setup}

The models were implemented using TensorFlow and Keras with GPU acceleration. The reported training configuration used a batch size of 32, the Adam optimizer, and categorical cross-entropy loss. ReduceLROnPlateau was used to reduce the learning rate when validation performance stopped improving, allowing the optimizer to make smaller parameter updates during later stages of training. Early stopping was applied when validation loss failed to improve. The models were trained for up to 20 epochs, subject to early stopping.

For MobileNetV2, the pre-trained feature-extraction layers were initially used with a task-specific classification head for the seven currency classes. The implementation also describes subsequent fine-tuning of the upper layers to further adapt the pre-trained model to the Nepali currency classification task.

\section{Results and Findings}

\subsection{Overall Performance}

\begin{table}[H]
\centering
\small
\caption{Performance comparison of the three model configurations.}
\label{tab:results}
\begin{tabular}{@{}lccccc@{}}
\toprule
\textbf{Model} & \textbf{Val.\ accuracy} & \textbf{Test accuracy} & \textbf{Precision} & \textbf{Recall} & \textbf{F1-score} \\
\midrule
Basic CNN                          & 87\% & 85\% & 0.83 & 0.81 & 0.82 \\
MobileNetV2 (transfer learning)    & 91\% & 89\% & 0.89 & 0.87 & 0.88 \\
MobileNetV2 (hyperparameter tuning)& 95\% & 84\% & 0.93 & 0.91 & 0.92 \\
\bottomrule
\end{tabular}
\end{table}

The baseline CNN achieved 87\% validation accuracy and 85\% test accuracy. MobileNetV2 with transfer learning achieved 91\% validation accuracy and 89\% test accuracy, showing an improvement over the baseline CNN on the reported independent test result.

The hyperparameter-tuned MobileNetV2 achieved the highest reported validation accuracy at 95\%. The reported precision, recall, and F1-score for this configuration were 0.93, 0.91, and 0.92, respectively. However, its reported test accuracy was 84\%, which was lower than the 89\% achieved by MobileNetV2 with transfer learning.

This difference shows that the model with the highest validation accuracy did not achieve the highest independent test accuracy. It also highlights the importance of considering independent test performance when comparing model configurations, rather than relying on validation accuracy alone.

\begin{figure}[H]
\centering
\includegraphics[width=0.85\textwidth]{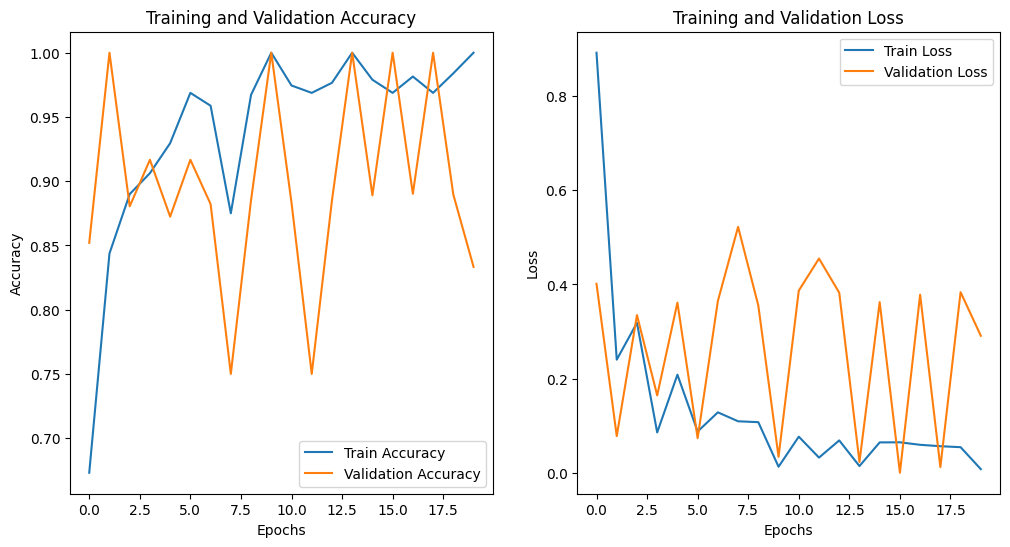}
\caption{Training and validation accuracy and loss curves for the baseline CNN.}
\label{fig:cnncurves}
\end{figure}

Figure~\ref{fig:cnncurves} shows the training and validation accuracy and loss curves for the baseline CNN. The training accuracy increases over the epochs, while the validation accuracy shows considerable fluctuation, indicating a difference between training and validation behaviour. The figure is presented as an example of the learning behaviour observed during the experiments rather than as a direct comparison of all three model configurations.

\subsection{Precision-Recall and ROC Analysis}

\begin{figure}[H]
\centering
\includegraphics[width=0.75\textwidth]{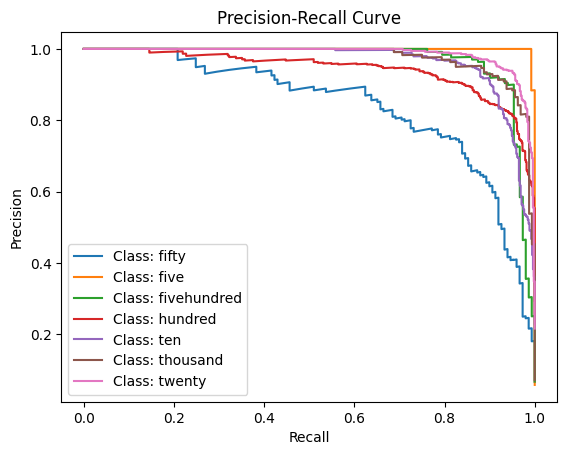}
\caption{Precision-recall curves for the seven Nepali currency denominations using the hyperparameter-tuned MobileNetV2 model.}
\label{fig:pr}
\end{figure}

Precision-recall analysis provides class-specific information about the relationship between precision and recall. The curve shown in Figure~\ref{fig:pr} contains separate curves for the seven denomination classes. The five, five-hundred, and thousand-rupee classes maintain relatively high precision across much of the recall range, while the fifty-rupee class shows a more noticeable decline in precision as recall increases. The fifty-rupee curve shows a more noticeable decline in precision as recall increases, which is consistent with the classification difficulties observed for this denomination in the error analysis.

\begin{figure}[H]
\centering
\includegraphics[width=0.85\textwidth]{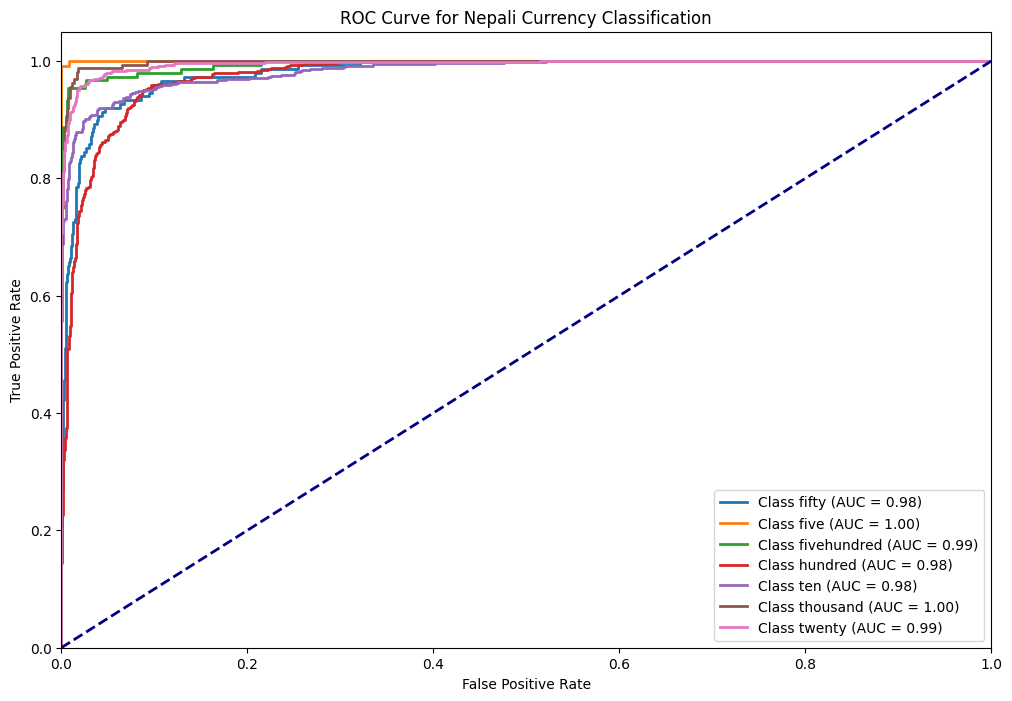}
\caption{ROC curves and class-wise AUC values for the hyperparameter-tuned MobileNetV2 model.}
\label{fig:roc}
\end{figure}

The class-wise ROC analysis shown in Figure~\ref{fig:roc} provides an additional view of class discrimination for the hyperparameter-tuned MobileNetV2 model. The reported AUC values are 0.98 for fifty, 1.00 for five, 0.99 for five hundred, 0.98 for hundred, 0.98 for ten, 1.00 for thousand, and 0.99 for twenty. These AUC values indicate strong class-separation behaviour within the evaluated dataset. However, they should be interpreted alongside the independent test accuracy because ROC-AUC and overall accuracy measure different aspects of classification performance.

\subsection{Web Implementation}

A web-based currency classification prototype was developed to demonstrate the practical use of the trained model. The backend was implemented using Flask, while web technologies were used to provide the user interface and interaction with the classifier.

The application provides two primary interaction modes. In the first mode, the user can capture or upload an image of a Nepali banknote. The image is processed through the classification pipeline, and the predicted denomination is provided through both text and voice output. In the second mode, the application provides a webcam interface where the user positions the banknote within the designated viewing area. The system processes the visual input and provides a voice announcement of the predicted denomination.

The prototype therefore demonstrates both conventional image-input classification and webcam-based interaction. Voice feedback also provides an additional accessibility-oriented output channel. The web prototype is presented as a practical demonstration of the classifier rather than as a separately benchmarked real-time system. The experiments reported in this study focus on classification performance, and no formal user study or systematic latency benchmark was conducted.

\begin{figure}[H]
\centering
\includegraphics[width=0.8\textwidth]{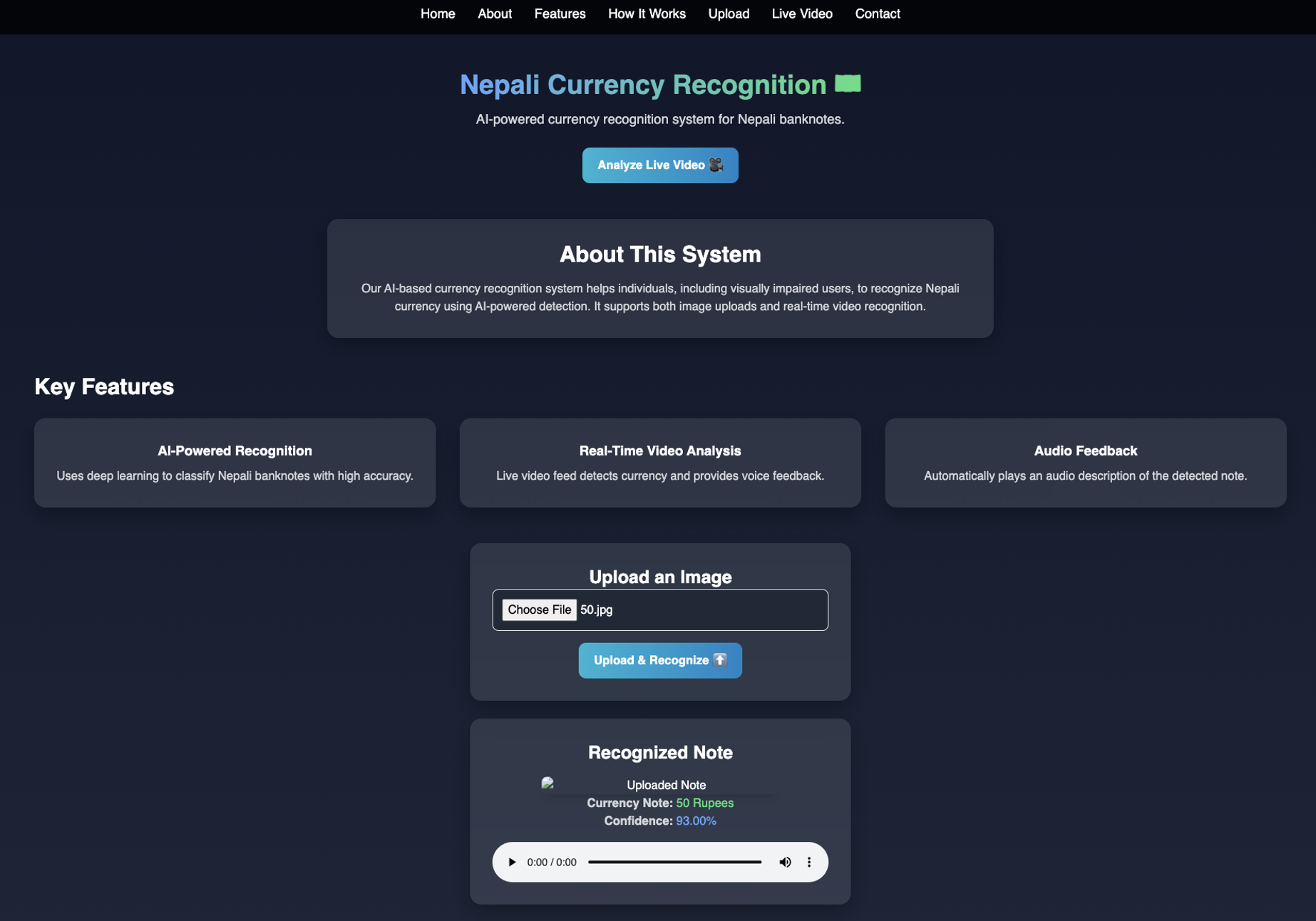}
\caption{Web-based interface for Nepali currency denomination classification.}
\label{fig:web}
\end{figure}

\section{Discussion}

\subsection{Interpretation of Results}

The experimental results show that the choice of model configuration affected classification performance. The baseline CNN achieved a reported test accuracy of 85\%, while MobileNetV2 with transfer learning achieved 89\%. Under the reported experimental conditions, this represents an improvement of four percentage points in test accuracy.

The hyperparameter-tuned MobileNetV2 achieved the highest validation accuracy at 95\%, along with the strongest reported precision, recall, and F1-score values. However, its reported test accuracy was 84\%, which was lower than the 89\% achieved by the untuned transfer-learning configuration. The difference between validation and test performance is therefore an important observation. The results indicate that increasing validation accuracy through hyperparameter tuning did not necessarily result in the highest held-out test accuracy in this experiment. Therefore, the tuned model should not be described as the highest-performing configuration across all metrics.

The class-level analysis also indicates that some denominations were more difficult to distinguish than others. The fifty- and hundred-rupee classes were identified as more prone to confusion, while several other denominations showed stronger classification behaviour.

Regarding Research question 1, the three evaluated configurations successfully classified the seven Nepali denominations, with the tuned MobileNetV2 achieving the highest validation accuracy of 95\%. However, its 84\% test accuracy was lower than the 89\% achieved by the untuned MobileNetV2, showing that the highest validation result did not translate directly into the best held-out test result.

Regarding Research question 2, the observed limitations included class imbalance in the test set, confusion between visually similar denominations, and sensitivity to variations in image quality and acquisition conditions.

\subsection{Error Analysis and Real-World Applicability}

\begin{figure}[H]
\centering
\includegraphics[width=0.75\textwidth]{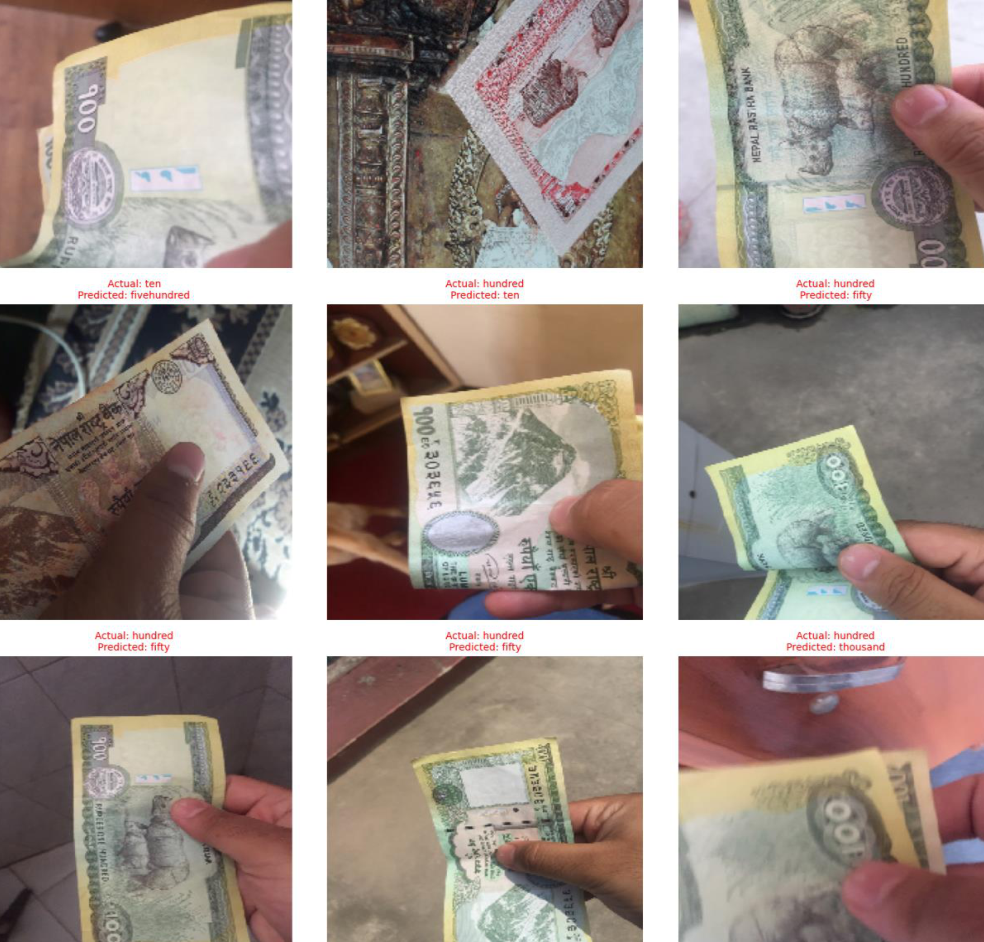}
\caption{Examples of misclassified Nepali banknote images from the evaluated test set.}
\label{fig:errors}
\end{figure}

The error examples shown in Figure~\ref{fig:errors} illustrate that visually similar banknotes can result in incorrect predictions. The original analysis identifies similarities in visual appearance, including colour and design characteristics, as possible contributors to these errors. Image quality is another relevant consideration. Poor lighting, blur, folds, and partial visibility can affect the visual information available to the classifier. Therefore, the test results should be interpreted as performance on the evaluated dataset rather than as a guarantee of performance under all uncontrolled environmental conditions.

The class distribution of the test dataset is also uneven, with the hundred- and ten-rupee classes representing substantially larger portions of the test set than several other denominations. This imbalance provides an additional reason to consider class-level precision, recall, F1-score, and ROC analysis alongside overall accuracy. Broader real-world evaluation would require independently collected images captured under uncontrolled conditions, including variations in lighting, backgrounds, viewing angles, camera quality, and different levels of banknote wear.

\subsection{Limitations}

Several limitations should be considered when interpreting the results. First, the study focuses on seven Nepali currency denominations and therefore does not establish performance for other currencies without additional training.

Second, the test dataset is not evenly distributed across the seven classes. Although the training and validation subsets contain equal numbers of images per class, the test subset contains unequal class counts. Therefore, class-level metrics and curves are important complements to the overall test accuracy.

Third, the dataset contains variation in image conditions, but the study does not provide a comprehensive controlled benchmark covering all possible conditions, including lighting, camera quality, background, banknote wear, blur, occlusion, and viewing angle.

Fourth, the reported results show a difference between validation and test performance for the hyperparameter-tuned MobileNetV2 configuration. The difference between the 95\% validation accuracy and 84\% held-out test accuracy indicates that validation performance alone was not sufficient to identify the configuration with the highest test accuracy.

Fifth, the present study addresses denomination classification rather than counterfeit authentication. A genuine-versus-counterfeit classifier would require a dedicated dataset and a separate evaluation.

Finally, although the web prototype includes voice-assisted output and webcam-based interaction, no formal user study was conducted to quantify accessibility benefits, and no systematic end-to-end latency benchmark was performed.

\section{Conclusion and Future Work}

This study investigated CNN-based approaches for classifying seven Nepali currency denominations using a compiled dataset containing 16,249 images. Three configurations were compared: a baseline CNN, MobileNetV2 with transfer learning, and MobileNetV2 with hyperparameter tuning.

The baseline CNN achieved 85\% test accuracy, while MobileNetV2 with transfer learning achieved the highest reported test accuracy of 89\%. The hyperparameter-tuned MobileNetV2 achieved the highest reported validation accuracy of 95\% and the reported precision, recall, and F1-score values of 0.93, 0.91, and 0.92, respectively. However, its reported test accuracy was 84\%. This difference shows that the highest validation score did not translate into the highest independent test accuracy.

The error analysis indicates that visually similar denominations can be more difficult to distinguish and that image quality is an important consideration for broader deployment. The class-wise ROC analysis also showed high reported AUC values across the seven denomination classes.

A Flask-based web prototype was developed to demonstrate the practical application of the classifier. The system provides image capture or upload with text and voice output, as well as a webcam-based interaction mode with voice feedback. These features demonstrate the potential for combining denomination classification with accessibility-oriented interaction.

Overall, the study provides an experimental comparison of CNN and MobileNetV2 approaches for Nepali currency denomination classification and establishes a basis for future research involving larger datasets, more diverse acquisition conditions, dedicated counterfeit detection, and broader accessibility evaluation.

Future work should first focus on expanding the dataset with independently collected images representing different lighting conditions, backgrounds, camera devices, viewing angles, levels of banknote wear, folds, blur, and partial visibility. Such data would allow the generalization of the classifier to be evaluated more rigorously.

A dedicated counterfeit-detection component can then be developed using appropriately labelled genuine and counterfeit banknotes. This should be treated as a separate classification task from denomination recognition and evaluated using its own dataset and metrics.

Further research can investigate deployment on mobile and edge devices using lightweight inference approaches. The voice-assisted component can also be evaluated through formal usability and accessibility studies involving representative users.

Explainability methods such as Grad-CAM or saliency-based approaches could be investigated to determine which visual regions contribute to denomination predictions.

\nocite{cs2022currency,giri2021examination,schmidt2020transparency}

\bibliographystyle{plainnat}
\bibliography{ref}

\end{document}